\documentclass[letterpaper]{article} 
\usepackage{aaai2026}  

\usepackage{times}  
\usepackage{helvet}  
\usepackage{courier}  
\usepackage[hyphens]{url}  
\usepackage{graphicx} 
\usepackage{natbib}  
\usepackage{caption} 
\usepackage{algorithm}
\usepackage{algorithmic}
\usepackage{amsmath}
\usepackage{newfloat}
\usepackage{listings}
\usepackage{booktabs}
\usepackage{multirow}
\usepackage{threeparttable}
\usepackage{graphicx}
\usepackage{subcaption}
\usepackage{float}
\usepackage{times}
\usepackage{helvet}
\usepackage{courier}
\usepackage{xcolor}

\DeclareCaptionStyle{ruled}{labelfont=normalfont,labelsep=colon,strut=off} 
\floatstyle{ruled}
\newfloat{listing}{tb}{lst}{}
\floatname{listing}{Listing}
\title{Multi-Plasticity Synergy with Adaptive Mechanism Assignment for Training Spiking Neural Networks}
\author {
    Yuzhe Liu\textsuperscript{\rm 1},
    Xin Deng\textsuperscript{\rm 2},
    Qiang Yu\textsuperscript{\rm 1*},
}
\affiliations {
    \textsuperscript{\rm 1}College of Intelligence and Computing, Tianjin University\\
    \textsuperscript{\rm 2}Chongqing University of Post and Telecommunications \\
    *Corresponding: yuqiang@tju.edu.cn
}

\usepackage{bibentry}

\begin{document}

\maketitle

\begin{abstract}
Spiking Neural Networks (SNNs) are promising brain-inspired models known for low power consumption and superior potential for temporal processing, but identifying suitable learning mechanisms remains a challenge. Despite the presence of multiple coexisting learning strategies in the brain, current SNN training methods typically rely on a single form of synaptic plasticity, which limits their adaptability and representational capability. In this paper, we propose a biologically inspired training framework that incorporates multiple synergistic plasticity mechanisms for more effective SNN training. Our method enables diverse learning algorithms to cooperatively modulate the accumulation of information, while allowing each mechanism to preserve its own relatively independent update dynamics. We evaluated our approach on both static image and dynamic neuromorphic datasets to demonstrate that our framework significantly improves performance and robustness compared to conventional learning mechanism models. This work provides a general and extensible foundation for developing more powerful SNNs guided by multi-strategy brain-inspired learning.
\end{abstract}


\section{Introduction}

Spiking neural networks (SNNs) have attracted increasing attention due to their event-driven nature, temporal processing capability, and energy efficiency \cite{roy2017programmable,kasabov2013dynamic,kim2020spiking,yu2020toward,yao2023attention}, making them a promising alternative to traditional artificial neural networks (ANNs). With their bio-inspired neuronal dynamics \cite{maass1997networks,ros2006event,deco2008dynamic}, SNNs exhibit significant potential to build brain-inspired computational models and deploy low-power neuromorphic systems. However, despite their advantages, training SNNs effectively remains a significant challenge. Most existing learning methods for SNNs—such as Spatio-temporal Backpropagation (STBP) \cite{wu2018spatio}, Backpropagation Through Time (BPTT) \cite{lee2016training}, Spike-Timing-Dependent Plasticity (STDP) \cite{kheradpisheh2018stdp}, self-backpropagation (SBP) \cite{zhang2021self}, or Hybrid global-local learning (HGLL) \cite{wu2022brain} typically adopt a single learning paradigm or merely combine two learning mechanisms. While these methods provide distinct advantages across various tasks, they arguably exhibit certain limitations in representing complex neuronal dynamics, extracting comprehensive features, and improving the model's generalization capability and robustness.

\begin{figure}[t]
    \centering
    \includegraphics[width=1\linewidth]{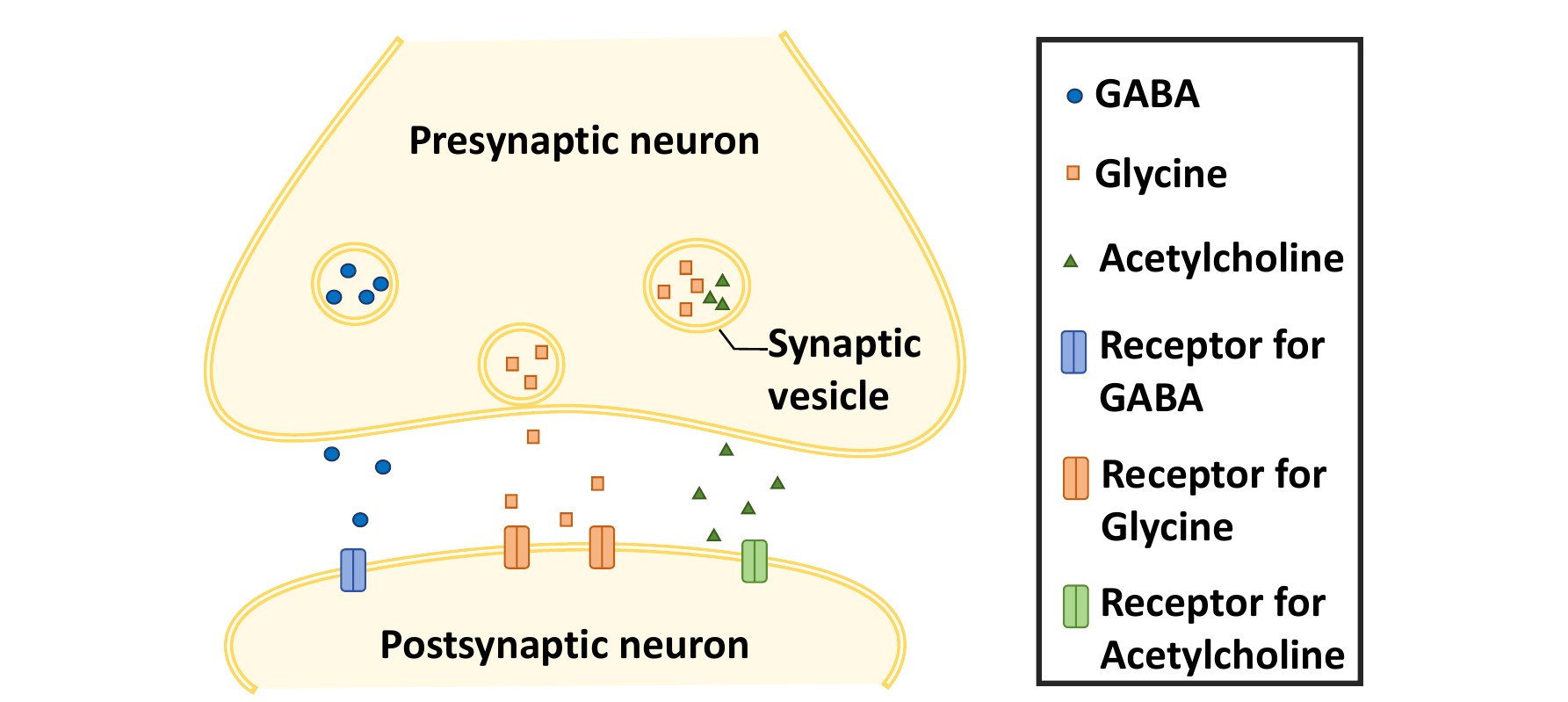}
    \caption{The presynaptic neuron releases multiple types of neurotransmitters via vesicles into the synaptic cleft, and these neurotransmitters subsequently bind to their respective receptors on the postsynaptic neuron. The quantities of released neurotransmitters and the corresponding receptors on the postsynaptic neuron vary across types. The neurotransmitter types in this figure are schematic representations, and the specific combinations vary depending on the cell type.  }
    \label{fig:neurotransmitter}
\end{figure}
In contrast, biological brains rely on the coordinated action of multiple learning mechanisms operating at different spatial and temporal scales \cite{kandel1991cellular,edelmann2017coexistence,gonzalez2025synaptic,el2019exercise}. Yet, the simultaneous optimization of multiple plasticity rules remains largely underexplored in computational models. Therefore, integrating these heterogeneous learning mechanisms within a unified computational framework offers a promising approach to harness their complementary strengths and synergistic effects across diverse learning paradigms. Nevertheless, enabling effective co-adaptation among these distinct mechanisms, while further incorporating the complex and diverse spiking dynamics, remains a significant challenge.

It has long been observed in neuroscience that individual neuron can co-release multiple neurotransmitters, each influencing synaptic activity through distinct molecular pathways \cite{jaim1994co,gutierrez2008pluribus,hnasko2012neurotransmitter,vaaga2014dual}. This phenomenon, known as neurotransmitter co-existence, reveals a fundamental principle of neural computation: multiple signaling processes can operate in parallel, maintaining functional independence while collectively achieving an integrated regulatory outcome \cite{bloomfield2001rod,tritsch2016mechanisms,rozycka2017space}. Figure \ref{fig:neurotransmitter} exemplifies this principle by presenting a conceptual framework wherein coexisting neurotransmitters regulate synaptic plasticity through distinct yet synchronized signaling pathways. This neuro-computational principle enables the brain to dynamically adapt to heterogeneous stimuli, providing stable yet flexible responses to ever-changing environments and forming the foundation for perception, cognition, and action \cite{lee2016segregated,liang2025excitation}.
    
Building on neuroscientific principles of signal synergy, we propose a Multiple Plasticity Synergy Learning (MPSL) framework that computationally emulate biological learning processes. Our method enables diverse learning mechanisms with complementary strengths to collaboratively guide the accumulation of membrane potential, while preserving their distinct update behaviors. Inspired by the observation that the quantities of co-released neurotransmitters and their corresponding receptors vary across neuron types \cite{hansen2022mapping}, we normalize these factors into a unified learnable parameter that adaptively modulates the contribution of each mechanism during the fusion process. This design aims to leverage the complementary strengths of different learning mechanisms, leading to improved performance, generalization, and robustness across a variety of tasks. The proposed framework maintains architectural generality to a broad range of learning mechanisms. For experimental validation, we adopt three representative learning mechanisms: STBP, Hebbian learning and SBP. 

Overall, our main contributions are threefold:
\begin{itemize}
    \item We propose a biologically inspired MPSL framework for spiking neural networks, motivated by the phenomenon of neurotransmitter co-existence. This framework enables multiple learning mechanisms to collaboratively guide the accumulation of information while preserving their distinct update behaviors.

    \item We introduce a learnable modulation parameter that adaptively balances the contributions of each learning mechanism during the fusion process. This design enables the network to flexibly adjust to varying learning demands, thereby significantly improving the framework’s flexibility and extensibility.
    
    \item 	We validate our approach on both static datasets (MNIST, Fashion MNIST, CIFAR-10) and dynamic ones (N-MNIST and DVS-Gesture), and further evaluate the network under noise and cropping conditions, demonstrating that our method effectively enhances both the performance and robustness of spiking neural networks.
    
\end{itemize}

\section{Related Works}
\paragraph{Supervised Learning in SNNs}
Supervised learning has been a critical driving force in the advancement of SNNs, enabling them to achieve competitive performance across a range of tasks. Unlike traditional ANNs which use continuous activation functions, SNNs rely on discrete spike events, introducing non-differentiable activation dynamics that challenge gradient-based optimization.  To address this, early works such as SpikeProp \cite{bohte2002error} and Tempotron \cite{gutig2006tempotron} proposed heuristic gradient-like update rules based on spike timing, providing foundational methods for supervised learning in the temporal domain. More recent methods adopt surrogate gradients, which approximate the non-differentiable spike function with smooth alternatives, thereby enabling BPTT. For example, STBP \cite{wu2018spatio} applies BPTT to SNNs by propagating gradients through both spatial layers and temporal membrane states, forming the backbone of modern supervised SNN training. Neftci et al.~\cite{neftci2019surrogate} further generalized this approach into a unified framework, establishing surrogate gradient learning as a scalable and biologically inspired solution for optimizing SNNs with standard backpropagation techniques. However, the dependence on explicit labels and global feedback limits their alignment with biological learning and applicability in unlabeled environments. These limitations highlight the need for complementary mechanisms.

\paragraph{Unsupervised Learning in SNNs}
Unsupervised learning plays a fundamental role in SNNs, often implemented through local synaptic plasticity based on the Hebbian principle of temporal correlation (neurons that fire together wire together) \cite{hebb1949organization}. Building on Hebb's seminal correlation principle, Oja \cite{oja1982simplified} introduced weight normalization to stabilize synaptic growth, albeit with reduced temporal sensitivity. Subsequent work explored temporal causality through STDP \cite{markram1997regulation, bi1998synaptic}, which leveraged precise pre-post spike intervals to enhance learning dynamics. Bienenstock \cite{bienenstock1982theory} further improved robustness by incorporating dynamic thresholds, though requiring manual parameter calibration. Gütig \cite{gutig2016spiking} extended these principles by integrating subthreshold membrane potentials, bridging discrete spiking events with continuous neuronal dynamics at the cost of implementation complexity. Mozafari \cite{mozafari2018bio} advanced task-driven adaptation through neuromodulatory signals, enabling more flexible plasticity rules. Despite their biological plausibility, these unsupervised methods often suffer from limited task adaptability. In this work, we incorporate Hebbian plasticity into a multi-mechanism learning framework to enhance temporal representation while preserving local learning characteristics.

\paragraph{Hybrid Learning in SNNs}
Grounded in the three-factor learning theory~\cite{fremaux2016neuromodulated, gerstner2018eligibility}---which integrates presynaptic activity, postsynaptic activity, and neuromodulatory signals---these approaches reinterpret global supervision (e.g., task-specific errors or rewards) as biologically plausible third factors modulating synaptic plasticity. Early theoretical foundations for plasticity optimization~\cite{bengio1992global} laid the groundwork for gradient-based supervision of local learning rules. Subsequent work extended these principles to non-spiking architectures, demonstrating broad applicability in meta-learning and dynamic adaptation \cite{munkhdalai2018metalearning, miconi2018differentiable}. Further innovations automated unsupervised rule  through meta-optimization~\cite{metz2018meta}, collectively advancing hybrid plasticity frameworks across learning paradigms. Wu~\cite{wu2022brain} integrated a brain-inspired Global-Local learning paradigm with a parametrized spiking differential dynamics model, facilitating efficient hybrid learning on neuromorphic hardware. In parallel, Zhang \cite{zhang2021self} established local feedback loops that approximate gradient descent without explicit error propagation, enabling coordinated synaptic modifications across neuron layers and significantly reducing computational costs. Their framework effectively integrates both STDP and SBP mechanisms, demonstrating applicability to both artificial and spiking neural networks. While prior hybrid approaches have explored integrating global and local learning rules, they are often limited to specific combinations or two-way interactions. In contrast, our work introduces a general framework that enables the coordinated use of multiple learning mechanisms, facilitating more flexible and biologically inspired training in SNNs. 

\begin{figure}[t]
    \centering
    \includegraphics[width=1\linewidth]{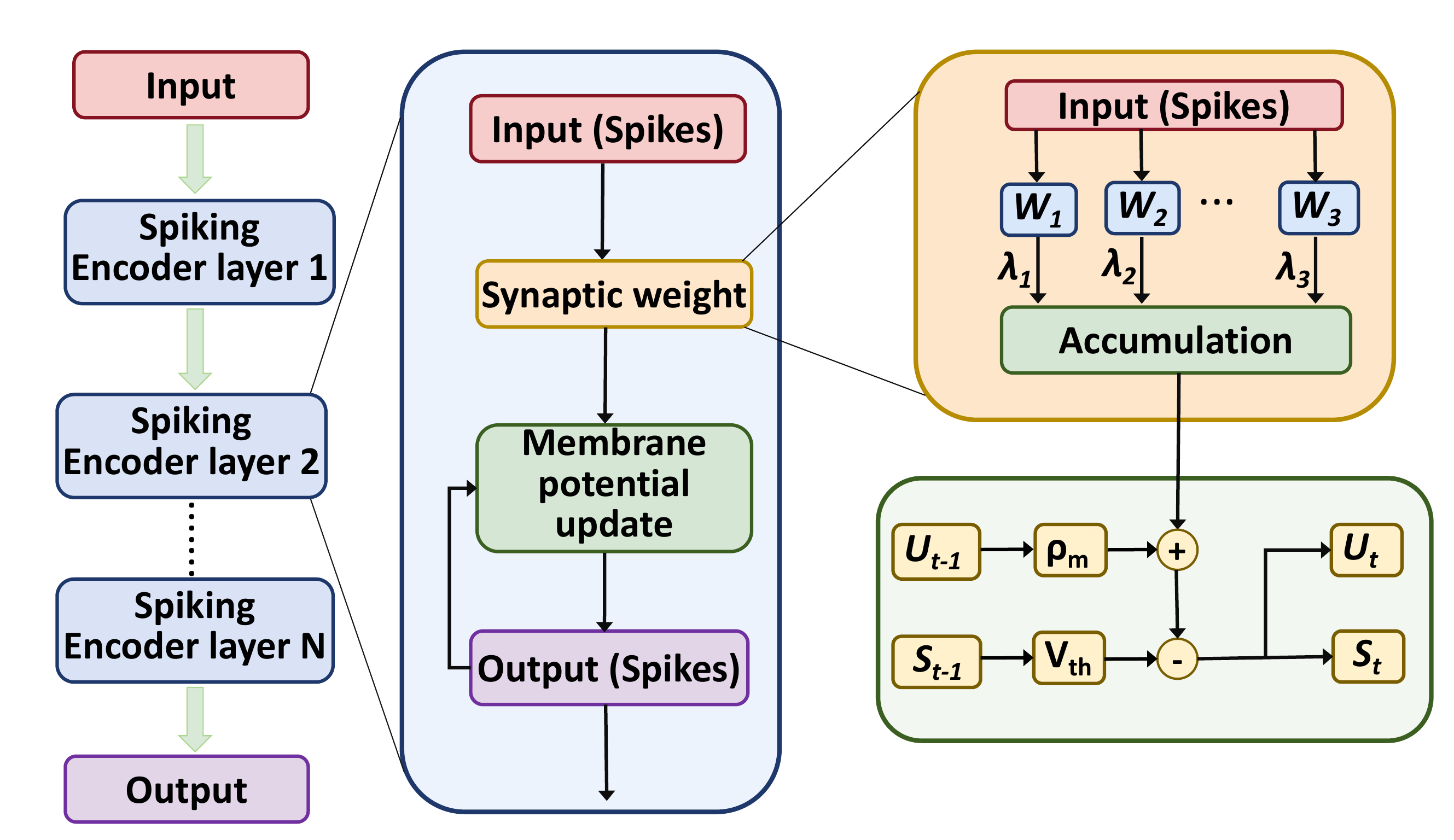}
    \caption{Overview of the MPSL training module and its internal submodules.}
    \label{fig:membrane update}
\end{figure}
\section{Methodology}
In this section, we introduce a biologically inspired MPSL training framework for SNN, which integrates multiple distinct learning algorithms into a unified computational process. We begin by revisiting the standard Leaky Integrate-and-Fire (LIF) neuron model and identifying the core equation that governs membrane potential updates. Based on this foundation, we detail how three learning mechanisms—STBP, Hebbian learning, and SBP—are incorporated and cooperatively contribute to the update dynamics.

We further present the interaction strategies among these mechanisms, highlighting how their interplay enables richer membrane potential modulation. We also analyze the parameter redundancy introduced by multi-path learning and provide a design rationale for its efficiency. Finally, we demonstrate the generality of our framework by emphasizing its compatibility with other learning rules and neural models.

\subsection{Spiking Neuron}

We begin with a detailed explanation of the LIF spiking neuron model and its iterative expression with the Euler method  
\cite{qiu2024gated}. The basis function of LIF neuron can be described as:
\begin{equation}
    S^{t, l}=\Theta\left(U^{t, l}-V_{\mathrm{th}}\right)=\left\{\begin{array}{ll}
1, & U^{t, l} \geq V_{\mathrm{th}} \\
0, & U^{t, l}<V_{\mathrm{th}}
\end{array}\right.,
\label{eq:spike}
\end{equation}
\begin{equation}
    I^{t,l}=f\left({~W}^{l}, S^{t, l-1}\right)
    \label{eq:input rule},
\end{equation}
\begin{equation}
    U^{t, l}=\rho_\mathrm{m} \left(U^{t-1, l}-S^{t-1, l} V_{\mathrm{th}}\right)+I^{t,l},
    \label{eq:membrane}
\end{equation}
where \(S^{t, l}\) and \(U^{t, l}\) represent the spike sequence and updated membrane potential at time step $t$ for $l$-th layer, respectively. Additionally, \(V_{\mathrm{th}}\) is the threshold that determines whether \(U^{t, l}\) results in a spike or remains silent. \({I}^{t,l}\) indicates the input at time step $t$ for $l$-th layer  and \(f\left(\cdot\right)\) is the function operation stands for convolution or fully connected computation. Moreover, \(W^{l}\) denotes synaptic weight matrix between two adjacent layers and \(\rho_\mathrm{m}\) is the membrane decay factor. To control the update of membrane potential, we adopt a soft reset strategy: when a spike is released, the membrane potential is subtracted by the threshold voltage.

Specifically, we refine Eq. \ref{eq:input rule} for MPSL learning compatibility. The original formulation, designed for a single update source, constrains integration of heterogeneous learning signals. Our reformulation enables concurrent contributions from multiple mechanisms (assuming there are \(n\) mechanisms). This yields coordinated modular updates—functionally analogous to neurotransmitter co-release—enhancing neuronal adaptability.
\begin{equation}
    {I}^{t,l}=\sum_{{i}=1}^{{n}}{f}\left({~W}_{i}^{l}, S^{t, l-1}\right).
\end{equation}

As is illustrated in Fig. \ref{fig:neurotransmitter}, variations in co-released neurotransmitter quantities and receptor distributions across neuron types motivate our normalization of these biological factors into adaptive scaling coefficients \(\lambda\) that dynamically modulate each mechanism's contribution during fusion. 
\begin{equation}
    {I}^{t,l}=\sum_{{i}=1}^{{n}} \lambda_{{i}} {f}\left({~W}_{i}^{l}, S^{t, l-1}\right).
\end{equation}

Figure \ref{fig:membrane update} presents our MPSL framework, where distinct computational principles (e.g., error-driven, correlation-based) differentially regulate neuronal behavior through specialized synaptic weight \({~W}_{i}\). Learnable parameter \(\lambda_{{i}}\) dynamically balances these influences, synthesizing their combined effect to guide membrane potential updates. This modular organization preserves each mechanism's unique dynamics while enabling synergistic optimization, ensuring both task adaptability and supporting flexible adaptation to diverse learning paradigms.

\subsection{Synaptic Plasticity}

While the proposed framework maintains architectural generality to accommodate diverse learning paradigms, we strategically implements three representative mechanisms (\(n\)=3): STBP (\({W}_{1}\)) for gradient-based temporal credit assignment, Hebbian plasticity (\({W}_{2}\)) for unsupervised correlation-driven adaptation and SBP (\({W}_{3}\)) for local-feedback gradient approximation. Next, we will introduce their respective update strategies.

As demonstrated by Eq.~\eqref{eq:spike}, the non-differentiable characteristics of spiking activity preclude direct application of backpropagation (BP) 
to SNNs. To address this fundamental limitation, 
surrogate gradient approaches replace the undefined gradients of spike generation with surrogate gradient functions \( u(\cdot) \) during error backpropagation. This methodology enables efficient training of SNNs 
through STBP.

Specifically, the SNN computes the spike activity via Eq.~\eqref{eq:spike} and its derivative with respect to the membrane potential through Eq.~\eqref{eq:surrogate}:
\begin{equation}
\frac{\partial S^{t, l}}{\partial U^{t, l}} \approx u’\left(U^{t, l}, V_{\mathrm{th}}\right),
\label{eq:surrogate}
\end{equation}

Following \cite{ding2024shrinking}, we simply adopt the rectangular function as the surrogate gradient in this paper:
\begin{equation}
u’\left(U^{t, l}, V_{\mathrm{th}}\right)=\frac{1}{{a}} \operatorname{sign}\left(\left|U^{t, l}-V_{\mathrm{th}}\right|<\frac{{a}}{2}\right),
\label{eq:surrogate function}
\end{equation}
where \(a\) serves as the hyperparameter controlling the shape of the rectangular function and we set it to 1. Through this way, chain rule-based weight updates for \({W}_{1}\) can be achieved.

We adopt a Hebbian-based variant from \cite{wu2022brain}, where synaptic weights \({W}_{2}\) are locally adjusted based on spike-timing correlation with a stabilizing decay term:
\begin{equation}
    W_{2}^{t,l}=W_{2}^{t-1,l}  e^{-\frac{\mathrm{dt}}{\tau_{\mathrm{w}}}}+\eta^{l} s^{t,l-1}\left(\rho\left(U^{t,l}\right)+\beta^{l}\right),
    \label{eq:hebbian rule}
\end{equation}
 where \({\mathrm{dt}}\) denotes the length of timestep, \(\tau_{\mathrm{w}}\) is the time constant, and \(\eta^{l}\) controls the local learning rate. \(\rho(\cdot)\) is a bounded nonlinear function and \(\beta^{l}\)  is an optional sliding threshold to control weight change directions.

The synaptic weights \({W}_{3}\) are trained using SBP \cite{zhang2021self}, where learning signals are generated locally without explicit gradient backpropagation. To align with our MPSL framework, we modify the original SBP by replacing its use of STDP updates with Hebbian-driven weight changes in \({W}_{2}\) as the learning signal. Specifically, Eq. \eqref{eq:sbp learning} and Eq. \eqref{eq:sbp update} detail the update rule of \({W}_{3}\). 
\begin{equation}
    W_{3}^{\mathrm{t}, l}=W_{3}^{t-1, l}e^{-\frac{d t}{\tau_{\mathrm{w}}}}+H(\Delta W_2^{t,l+1},\Delta W_2^{t,l}),
    \label{eq:sbp learning}
\end{equation}
\begin{equation}
    H(A,B) = \left(\lambda_{\mathrm{f}} E_{\text {diag }}\left(V+\lambda_{\mathrm{p}} \delta_{n}\left(R(A)\right)\right)\right)\cdot B,
    \label{eq:sbp update}
\end{equation}
\begin{equation}
    \delta_{n}(x)=\frac{x}{\sum_{i}^n x_{i}},
    \label{eq:sbp sigmax}
\end{equation}
where \(\lambda_{\mathrm{f}}, \lambda_{\mathrm{p}} \in [0.1, 1]\) are fraction factors, and \(E_{\text {diag }}(x)\) denotes the operation that maps a vector $x$ to a diagonal matrix with $x$ on its diagonal. Then, \(V\) is an all-ones vector and \(R(\cdot)\) here means sum of rows in a matrix.
 
To elucidate multi-plasticity update scheduling during training, we provide the framework's pseudocode specification in Algorithm 1. This specification precisely sequences the update timing for each learning rule across forward and backward passes, explicitly defining the application logic for individual rules, coordination protocols between optimization processes, and temporal orchestration of rule updates.

\begin{algorithm}[tb]
\caption{MPSL training framework}
\label{alg:algorithm}
\begin{algorithmic}[1] 
\REQUIRE input \(X\),  label \(Y\), time windows \(T\), number of network layers  \(L\), model weight \(W_1\), \(W_2\), \(W_3\).
\ENSURE updated network parameters.
\STATE Initialize membrane potential \(U\) and spiking output \(S\).
\FOR{$t=1$ to \(T\)}
\STATE $\boldsymbol{I}_1=\boldsymbol{X}$;
\FOR{$l=1$ to \(L\)}
\STATE \(U^{t,l}\) $\leftarrow$ Eq. {\ref{eq:input rule}, \ref{eq:membrane}}; // calculate membrane potential
\STATE \( S^{t,l}\) $\leftarrow$ Eq. \ref{eq:spike}; // calculate spiking output
\STATE \( W_2^{t,l}\) $\leftarrow$ Eq. \ref{eq:hebbian rule}; // update model weight \(W_2\)
\ENDFOR
\FOR{$l=L$ to 1}
\STATE \( W_3^{t,l}\) $\leftarrow$ Eq. {\ref{eq:sbp learning}, \ref{eq:sbp update}, \ref{eq:sbp sigmax}}; // update model weight \(W_3\)
\ENDFOR
\ENDFOR
\STATE \(O=S^L\);   
\STATE $\boldsymbol{\mathcal{L}}=\boldsymbol{\mathcal{L
}}(Y,O)$; // calculate loss
\STATE Calculate the gradient according to Eq. {\ref{eq:surrogate}, \ref{eq:surrogate function}}.
\STATE Update parameters $\boldsymbol{W_1}$ and other learnable parameters based on the STBP algorithm.
\end{algorithmic}
\end{algorithm}

\subsection{Mechanism Interaction}

Our method not only accommodates multiple learning paradigms but also enables their interactions to promote unified and efficient learning. Specifically, the global error signals from STBP are leveraged not only for updating its own synaptic weights but also for guiding parameter optimization in Hebbian learning (\(\eta^l,\beta^l\) in Eq. \eqref{eq:hebbian rule}) and SBP(\(\lambda_\mathrm{f},\lambda_\mathrm{p}\) in Eq. \eqref{eq:sbp update}). This introduces a level of indirect supervision that enhances the convergence and stability of the biologically inspired components.

Moreover,  SBP utilizes the synaptic update signals derived from Hebbian learning to perform its forward signal propagation. This reuse of Hebbian-modulated weight changes reflects a biologically grounded form of inter-process coordination, where plasticity in one pathway informs activity in another.

Finally, all these mechanisms contribute collectively to membrane potential dynamics. By co-modulating the potential updates through their respective signals, they form a coherent and synergistic learning process. This collaborative influence provides richer temporal dynamics and greater representational ability, while also improving the overall learning capacity of the network.
\begin{table*}[h]
\centering
\begin{tabular}{ccccccc}\toprule

Parameters& Descriptions& MNIST& FMNIST& NMNIST& CIFAR10& DVS-Gesture
\\\midrule 
Batch size& -& 100& 100& 100& 50& 32
\\
\(T\)& Time steps& 8& 8& 8& 8& 8
\\
\({\tau_{\mathrm{w}}}\)&  Time constant& 40& 40& 40& 200& 200\\ 
 \(V_{\mathrm{th}}\)& Threshold& 0.3& 0.4& 0.4& 0.5&0.4
\\
 \(N\)& Training epochs& 100& 100& 100& 150&150\\ \bottomrule
\end{tabular}
\caption{Parameter settings on different learning tasks.}
\label{tab:parameter settings}
\end{table*}

\subsection{Integration Rationale}

The proposed MPSL framework is built to leverage the complementary properties of different learning paradigms. STBP contributes strong global optimization capability by propagating gradients across time and layers. In contrast, Hebbian learning offers local plasticity through activity-dependent weight updates, which enhances adaptability and biological plausibility. Meanwhile, SBP introduces a form of self-consistency, where local signals are reused to reinforce parameter updates without relying on external labels or global loss. 

By integrating these mechanisms, the framework supports an interactive training process in which global supervision, local plasticity, and internal consistency collectively regulate synaptic changes. This interaction enhances the modulation of membrane potentials during learning and leads to improved temporal feature extraction. As a result, the model demonstrates stronger generalization and resilience, especially under sparse or noisy spiking conditions.

\subsection{Parameter Design}

To support multiple learning mechanisms, our framework assigns a dedicated weight component to each mechanism, allowing their respective plasticity rules to operate in parallel during the training process. These components are modulated through a shared coordination strategy that learns how to adaptively combine their contributions. Crucially, the individual weights are only maintained during training. For inference, they are linearly merged into a single effective weight using learned coefficients:
\begin{equation}
    {W}^{l}=\sum_{{i}=1}^{{n}} \lambda_{{i}} {W}_{i}^{l}.
\end{equation}
This approach ensures that the deployed model maintains the same parameter footprint as conventional single-rule models, introducing no extra cost. By decoupling the optimization process from the final model structure, the network can benefit from richer and more diverse learning signals during training without compromising runtime efficiency.

\subsection{Framework Generalization}

Although we instantiate our framework with three representative learning mechanisms—STBP, Hebbian learning, and SBP—the proposed design is not limited to this specific combination. Our core idea is to unify diverse learning rules under a shared membrane potential update process, while preserving the distinct learning pathways for each mechanism. As long as a learning algorithm can be formulated with local or global update signals and mapped to a membrane potential modulation, it can be naturally integrated into our framework. This includes, but is not limited to, other biologically inspired learning rules such as STDP, reward-modulated Hebbian plasticity, or other meta-learning strategies. This flexibility highlights the extensibility of our approach and its potential as a general foundation for future multi-mechanism SNN research.

\section{Experiments}

\subsection{Experimental Settings }

\subsubsection{Datasets.} We evaluate our proposed MPSL method on both static and dynamic spiking datasets to verify its generality and effectiveness across diverse data modalities.
For static image classification, we use MNIST, Fashion-MNIST, and CIFAR-10, which range from simple grayscale digits to complex natural scenes. For dynamic event-based recognition, we adopt N-MNIST and DVS-Gesture, two widely used benchmarks collected by neuromorphic vision sensors. These datasets present unique temporal structures and challenge models to capture fine-grained spatiotemporal patterns.

\subsubsection{Implementation Details. } All experiments are implemented using PyTorch (version 2.0.1) with CUDA 11.8 and conducted on a server equipped with NVIDIA RTX 4090 GPUs. We follow consistent training settings across all datasets unless otherwise stated. The details of the simulation time steps, training epochs, batch size, and some neuronal hyperparameters are summarized in Tab. \ref{tab:parameter settings}.

Moreover, we adopt the same network structures as in the prior work~\cite{wu2022brain} to ensure fair comparisons. For the MNIST, Fashion-MNIST and N-MNIST datasets, we use a five-layer CNN with the structure: [input--128C3--AP2--256C3--AP2--256C3--AP2--512FC--10]. For the DVS-Gesture dataset, we employ a nine-layer CNN: [input--64C3S2--BN--128C3S1--BN--256C3S1--BN--256C3S1--BN--256C3S1--BN--256C3S1--AP2--800FC--512FC--11FC]. We applied the batch normalization (BN) technique to convolutional layers on the DVS-Gesture dataset by following the work~\cite{fang2021incorporating}. For the CIFAR-10 dataset, we adopt a nine-layer CNN following the CIFARNet structure~\cite{wu2019direct}.

\begin{table*}[!tb]
 \centering
 \renewcommand{\arraystretch}{1.3}
 \begin{threeparttable}
 \begin{tabular}{clcccc}
  \toprule
 Dataset & Method & Architecture & Average timestep & Accuracy (\%)\\
  \midrule
  \multirow{5}{*}{MNIST} 
  &STDP \cite{diehl2015unsupervised} & Spiking MLP & 350\tnote{*} & 95.00 \\
  &SBP \cite{zhang2021self} & Spiking MLP & 10 & 95.14 \\
  &STBP \cite{wu2018spatio} & Spiking MLP & 30 & 98.89 \\
  &HGLL \cite{wu2022brain} & Spiking CNN & 8 & 99.50 \\
  &\textbf{MPSL} (Ours)  & Spiking CNN & 8 & \textbf{99.52} \\
  \hline
  \multirow{3}{*}{F-MNIST}
  &STBP \cite{wu2018spatio} & Spiking CNN\tnote{$\dag$} & 8 & 90.13 \\
  &HGLL \cite{wu2022brain} & Spiking CNN & 8 & 93.29 \\
   &\textbf{MPSL} (Ours)  & Spiking CNN & 8 & \textbf{93.72} \\
   \hline
   \multirow{3}{*}{CIFAR10}
   &STBP \cite{wu2018spatio} & Spiking CNN\tnote{$\dag$} & 8 & 90.55\\
   &HGLL \cite{wu2022brain} & Spiking CNN & 8 & 91.08 \\
   &\textbf{MPSL} (Ours)  & Spiking CNN & 8 & \textbf{92.12} \\
   \hline
   \multirow{3}{*}{N-MNIST}
   &STBP \cite{wu2018spatio} & Spiking CNN\tnote{$\dag$} & 8 & 98.78\\
   &HGLL \cite{wu2022brain} & Spiking CNN\tnote{$\dag$} & 8 & 99.25 \\
   &\textbf{MPSL} (Ours)  & Spiking CNN & 8 & \textbf{99.37} \\
   \hline
   \multirow{4}{*}{DvsGesture}
   &SBP \cite{zhang2021self} & Spiking MLP & 10 & 84.76 \\
   &STBP \cite{wu2018spatio} & Spiking CNN & 8 & 96.21\\
   &HGLL \cite{wu2022brain} & Spiking CNN & 8 & 97.01 \\
   &\textbf{MPSL} (Ours)  & Spiking CNN & 8 & \textbf{97.22} \\
  \bottomrule
 \end{tabular}
\end{threeparttable}
\caption{
Comparison with existing methods. * indicates average timestep calculated based on 1ms per step over a 350ms input duration, consistent with the poisson encoding in \cite{diehl2015unsupervised}. $\dag$ denotes results reproduced under the same conditions based on the authors’ released code.}
 \label{comparative}
\end{table*}

\subsection{Experimental Results}

We conduct extensive experiments to validate the effectiveness and robustness of our proposed MPSL method. The evaluation is carried out from three perspectives: accuracy comparison with existing methods, robustness under noise perturbations, and robustness under cropping perturbations.

\subsubsection{Accuracy Comparison.} We compare the classification performance of our proposed MPSL method with several representative learning algorithms for SNNs, including STDP, SBP, STBP and HGLL learning methods. Table~\ref{comparative} presents the accuracy results across both static (MNIST, Fashion-MNIST, CIFAR-10) and dynamic (N-MNIST, DVS-Gesture) datasets. Our approach consistently achieves top performance, reaching 99.52\% on MNIST, 93.72\% on Fashion-MNIST, 92.12\% on CIFAR-10, 99.37\% on N-MNIST, and 97.22\% on DVS-Gesture. Our method demonstrates overall superior generalization across diverse input modalities. These results validate the effectiveness of integrating complementary learning mechanisms and highlight the robustness of our method under various data conditions.

\subsubsection{Noise Robustness.} To evaluate the robustness of our method under input perturbations, we conduct inference with two types of commonly studied noise: Gaussian noise, which adds continuous-valued fluctuations to the input intensities, and salt-and-pepper noise, which randomly corrupts a proportion of pixels to either minimum or maximum intensity, simulating spike-like disturbances. Notably, both types of noise are applied under a same feedforward architecture to isolate the influence of the architecture. As is shown in Fig. ~\ref{fig:gaussian} and~\ref{fig:sp}, our proposed method consistently demonstrates superior accuracy under increasing noise levels compared to baseline models. The results suggest that integrating complementary learning rules improves generalization and enhances resilience to input corruption.

\subsubsection{Cropping Robustness.}
\begin{figure*}[t]
    \centering

    \begin{subfigure}[b]{0.32\textwidth}
        \includegraphics[width=\linewidth]{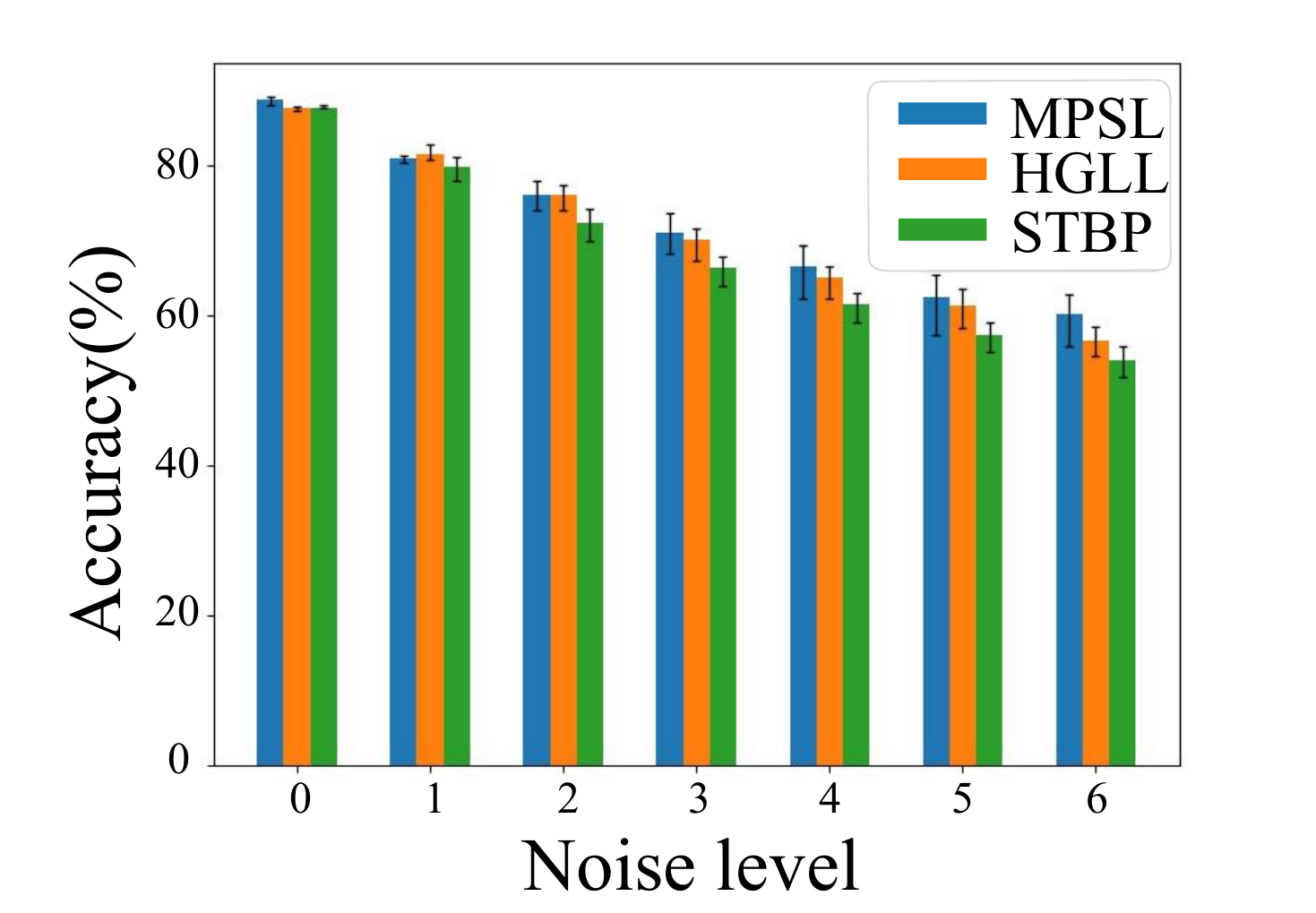}
        \caption{Gaussian noise}
        \label{fig:gaussian}
    \end{subfigure}
    \hfill
    \begin{subfigure}[b]{0.32\textwidth}
        \includegraphics[width=\linewidth]{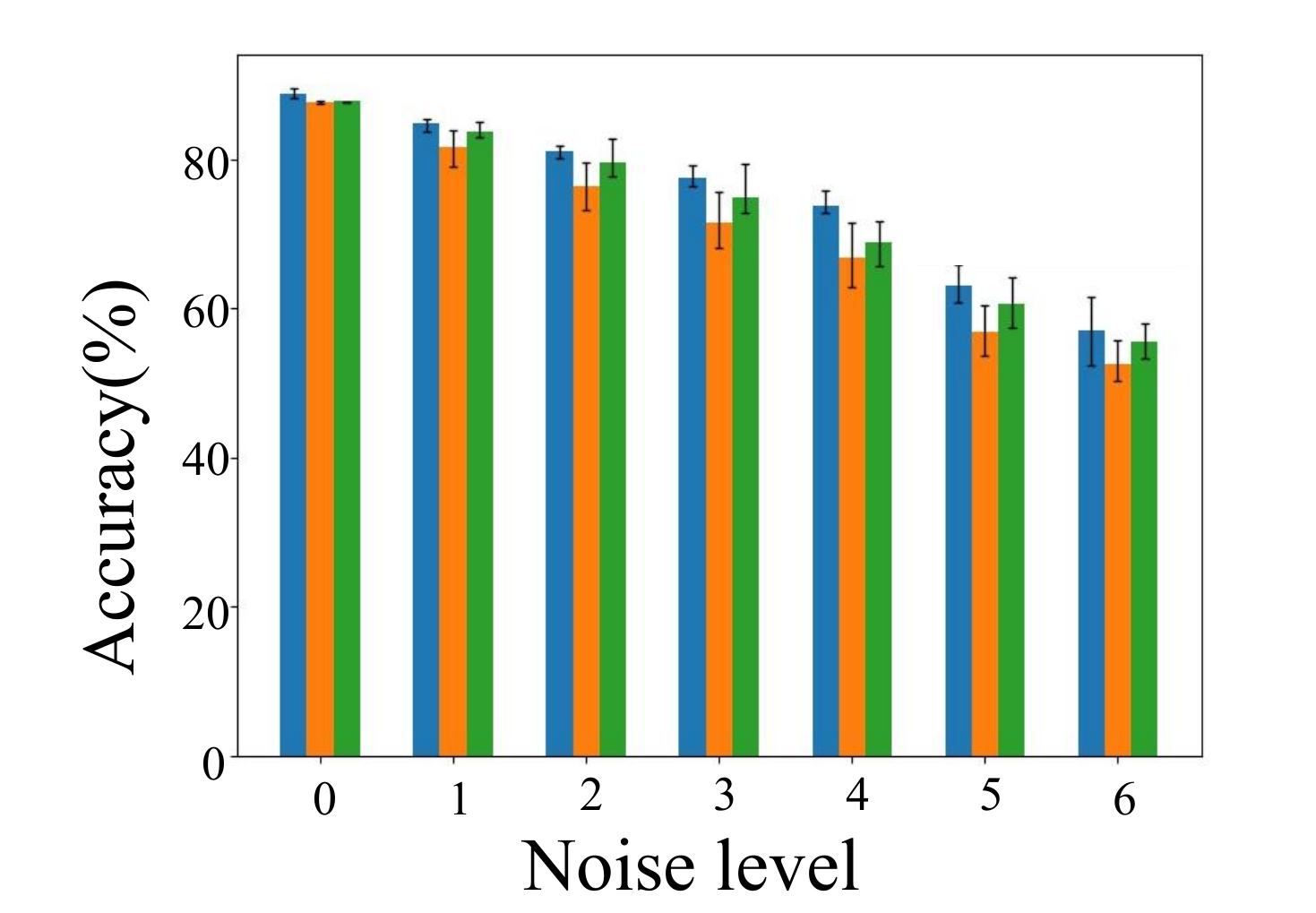}
        \caption{Salt-and-pepper noise}
        \label{fig:sp}
    \end{subfigure}
    \hfill
    \begin{subfigure}[b]{0.32\textwidth}
        \includegraphics[width=\linewidth]{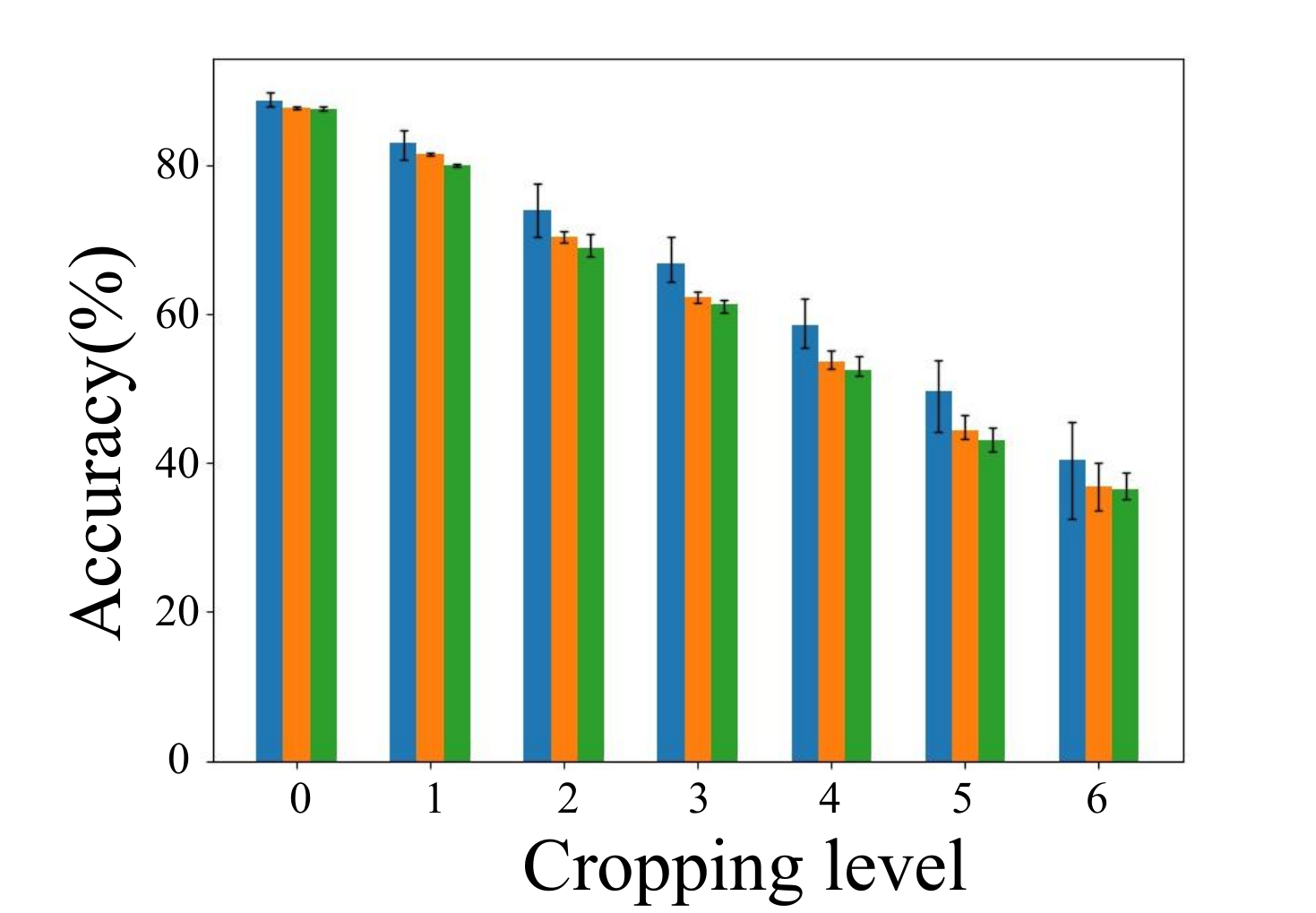}
        \caption{Cropping}
        \label{fig:cropping}
    \end{subfigure}

    \caption{Performance under varying levels of Gaussian noise, salt-and-pepper noise, and input cropping, where levels correspond to increasing variance for Gaussian noise, increasing noise amount for salt-and-pepper noise, and decreasing crop size for input cropping. Results are averaged over 5 runs and the error bars represent performance variability across multiple runs.  }
    \label{fig:noise-examples}
\end{figure*}

To further evaluate the resilience of the model to partial occlusion or information loss, we apply center cropping to input images with varying levels of severity \cite{hendrycks2019benchmarking,djolonga2021robustness}. Specifically, the original  images are cropped to square smaller central patches of size, simulating increasing degrees of spatial information reduction. As is shown in Fig. ~\ref{fig:cropping}, our method consistently outperforms baseline models across all cropping levels, maintaining significantly higher accuracy even under severe information loss. These results highlight the enhanced spatial redundancy and robustness of the multi-mechanism framework, which can better retain discriminative features despite partial input loss.
\subsection{Ablation Study}

\subsubsection{Effect of Learnable Fusion Coefficients.} To further investigate the importance of dynamically tuning the fusion coefficients that govern the interaction strength among learning mechanisms, we perform an ablation study under three settings:
(1) \textbf{Fixed Coefficients}: Equal static weights are assigned to each mechanism throughout training;
(2) \textbf{Learnable Coefficients}: The fusion coefficients are jointly optimized with the network parameters;
(3) \textbf{Frozen Learned Initialization}: Fusion coefficients are first trained in a learnable setting, then fixed during subsequent training.

\begin{figure}[t]
    \centering
    \begin{subfigure}[b]{0.48\linewidth}
        \includegraphics[width=\linewidth]{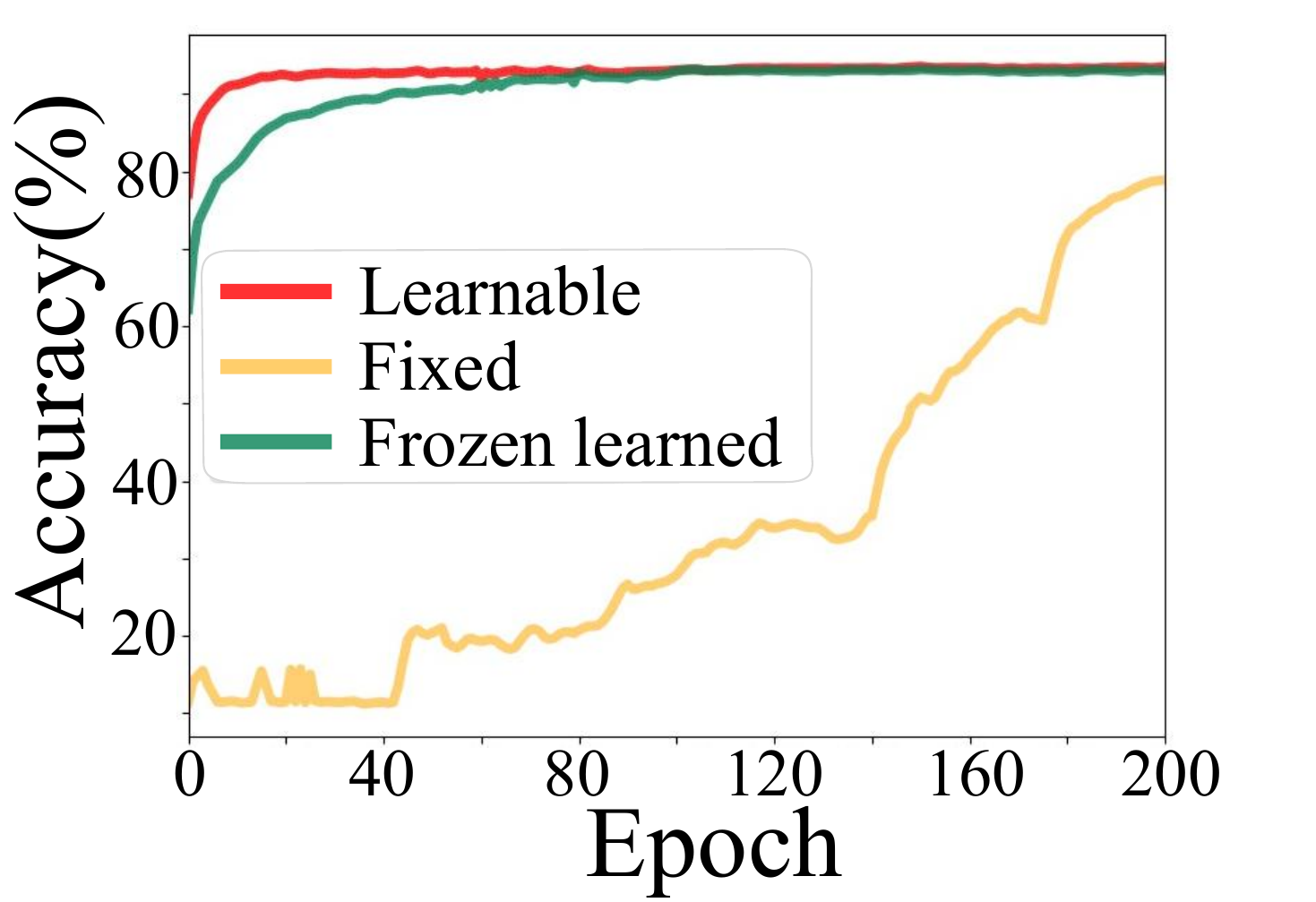}
        \caption{ Compare accuracy}
        \label{fig:b1}
    \end{subfigure}
    \hfill
    \begin{subfigure}[b]{0.48\linewidth}
        \includegraphics[width=\linewidth]{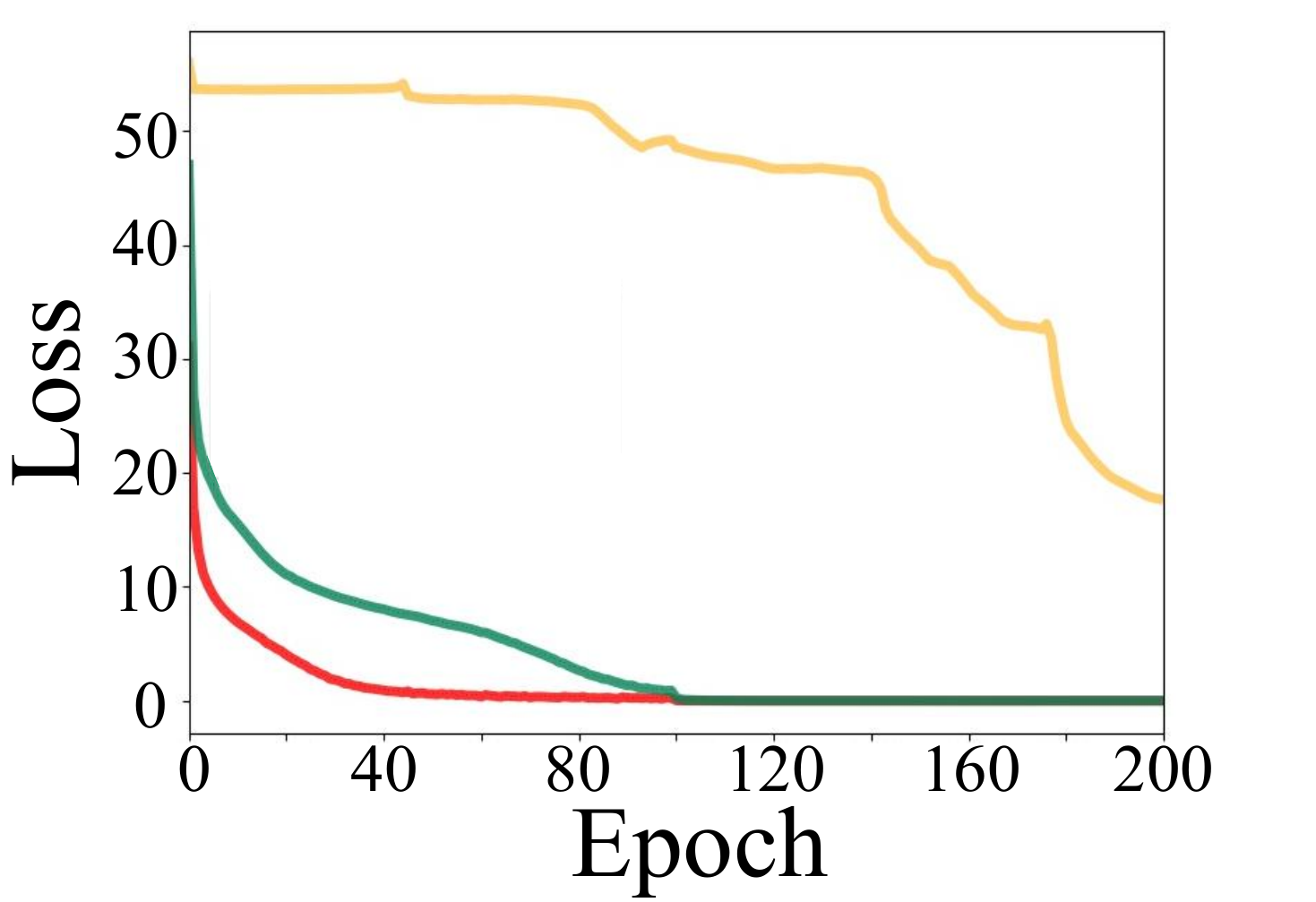}
        \caption{ Compare  loss}
        \label{fig:b2}
    \end{subfigure}
    \caption{Comparison of training accuracy and loss under different fusion coefficient strategies: fixed, learnable, and frozen learned. Results reflect the effect of fusion adaptability on model performance.}
    \label{fig:compare lamda}
\end{figure}

As is illustrated in Fig.~\ref{fig:compare lamda}, the learnable setting yields the highest performance, indicating that continuous adaptation of mechanism contribution is vital for aligning with task demands and input statistics. Notably, the frozen learned setting outperforms the fully fixed one, implying that the learned coefficients capture a favorable inductive bias. These results suggest that both the flexibility and the learned initialization of the fusion weights play a key role in maximizing the synergy among learning mechanisms.
\subsubsection{Feature Visualization.} To gain a deeper understanding of how different learning mechanisms affect the internal representations of the spiking neural network, we conduct a feature visualization experiment using t-SNE. We select 500 test samples from the Fashion-MNIST dataset and extract the membrane potentials from the penultimate layer, resulting in 128-dimensional feature vectors. These features are then projected into a two-dimensional space for visualization.

\begin{figure}[t]
    \centering
    \begin{subfigure}[b]{0.48\linewidth}
        \includegraphics[width=\linewidth]{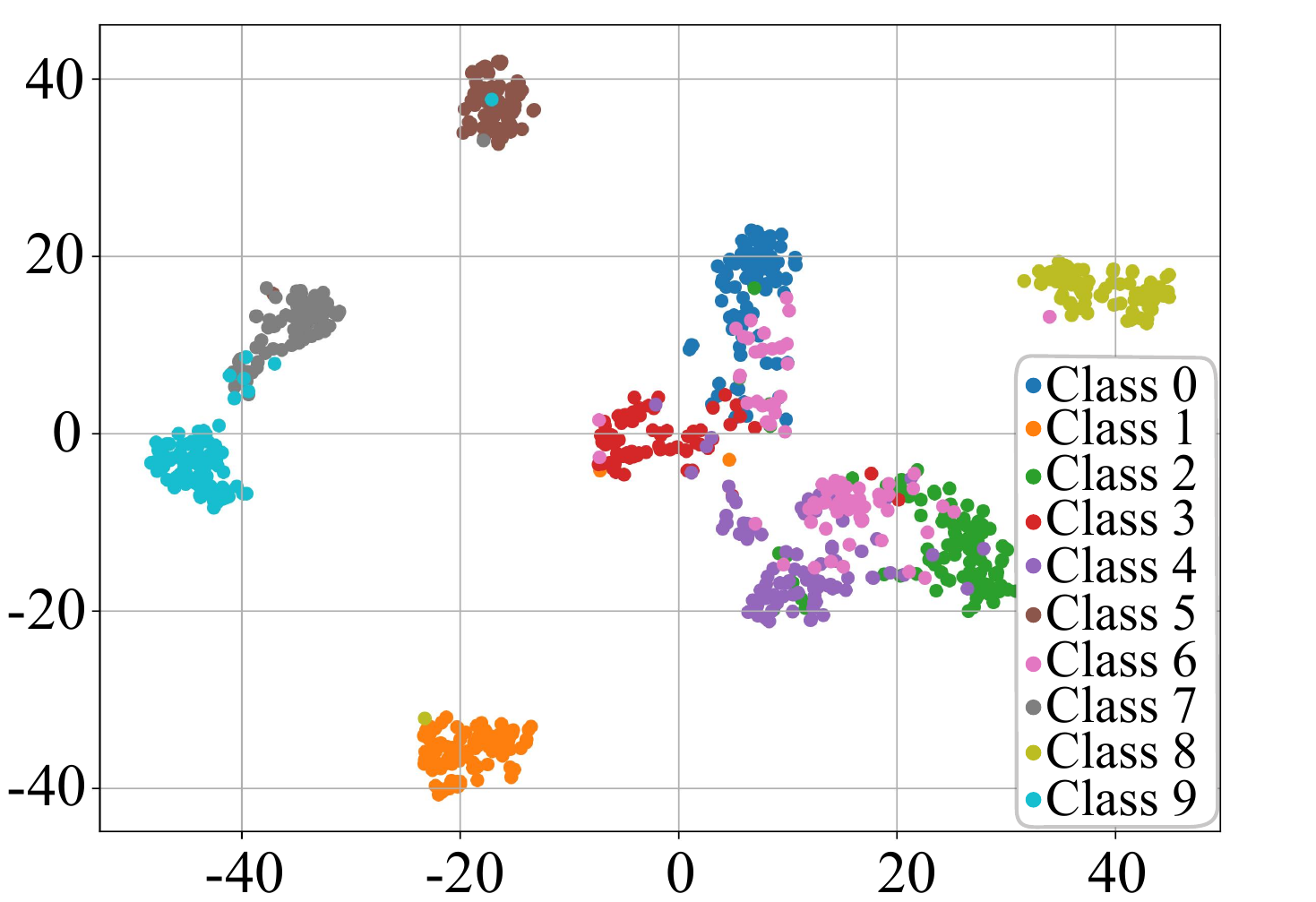}
        \caption{MPSL learning }
        \label{fig:c1}
    \end{subfigure}
    \hfill
    \begin{subfigure}[b]{0.48\linewidth}
        \includegraphics[width=\linewidth]{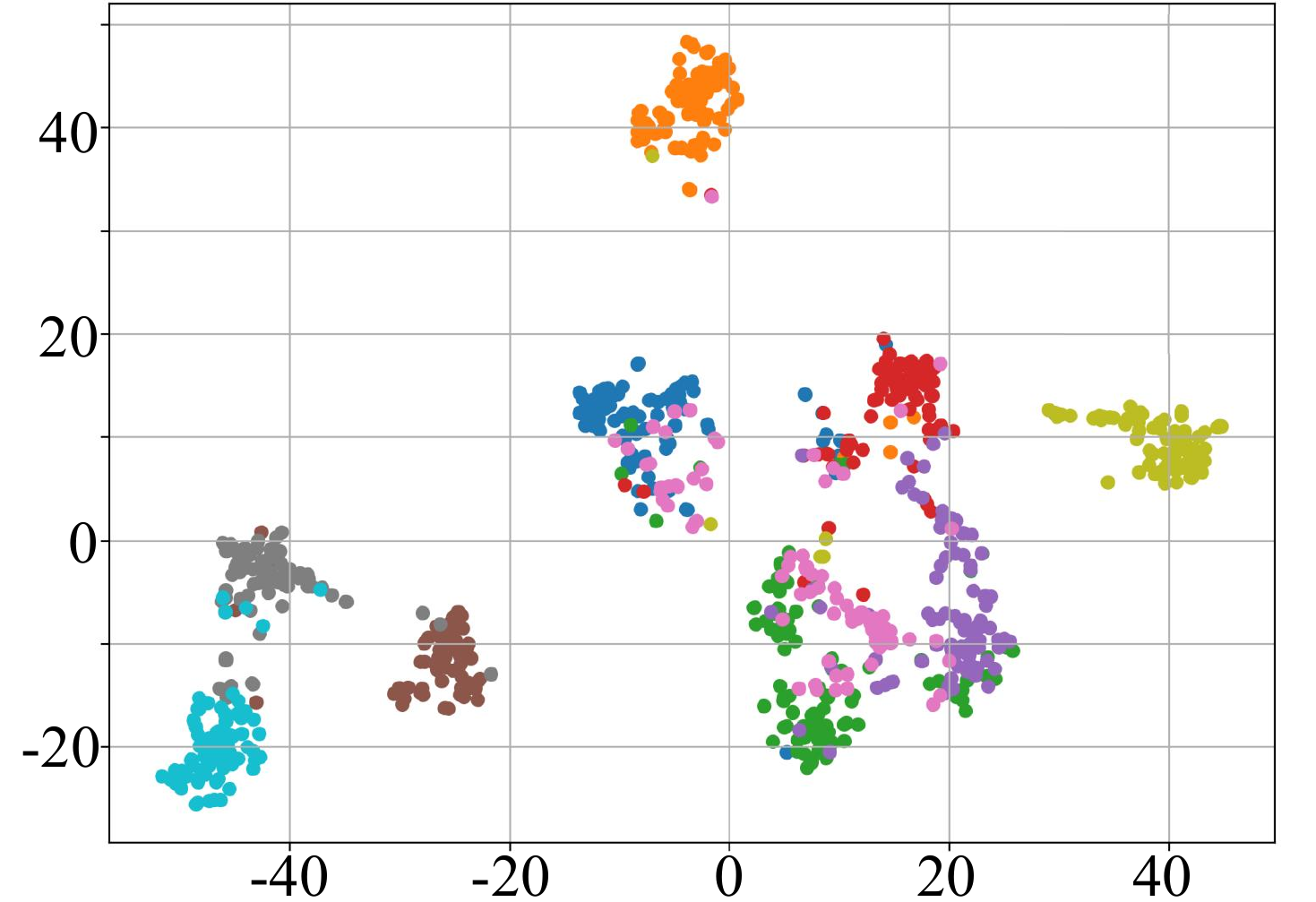}
        \caption{Without SBP}
        \label{fig:c2}
    \end{subfigure}
    
    \vspace{0.5em}

    \begin{subfigure}[b]{0.48\linewidth}
        \includegraphics[width=\linewidth]{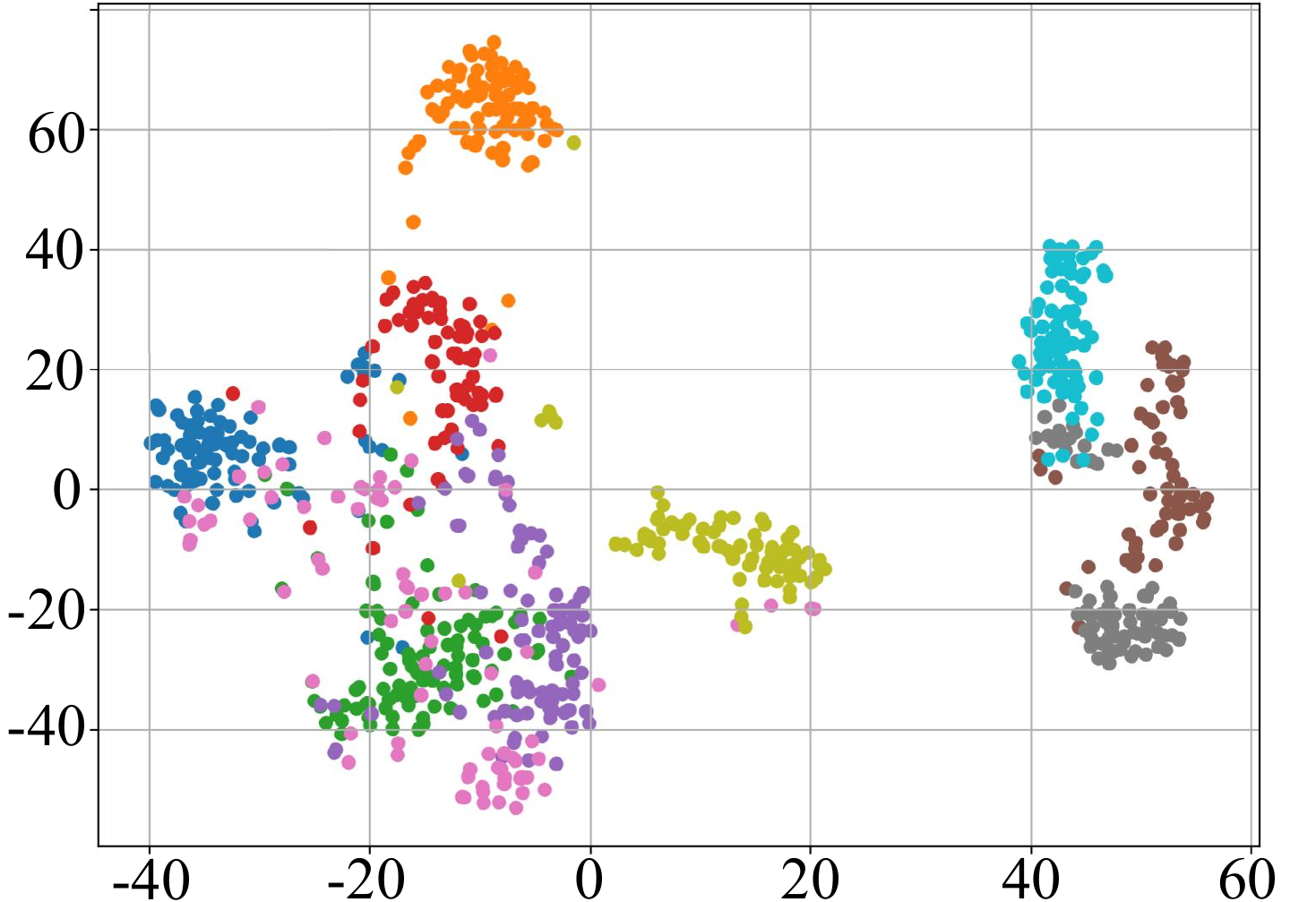}
        \caption{Without Hebbian learning}
        \label{fig:c3}
    \end{subfigure}
    \hfill
    \begin{subfigure}[b]{0.48\linewidth}
        \includegraphics[width=\linewidth]{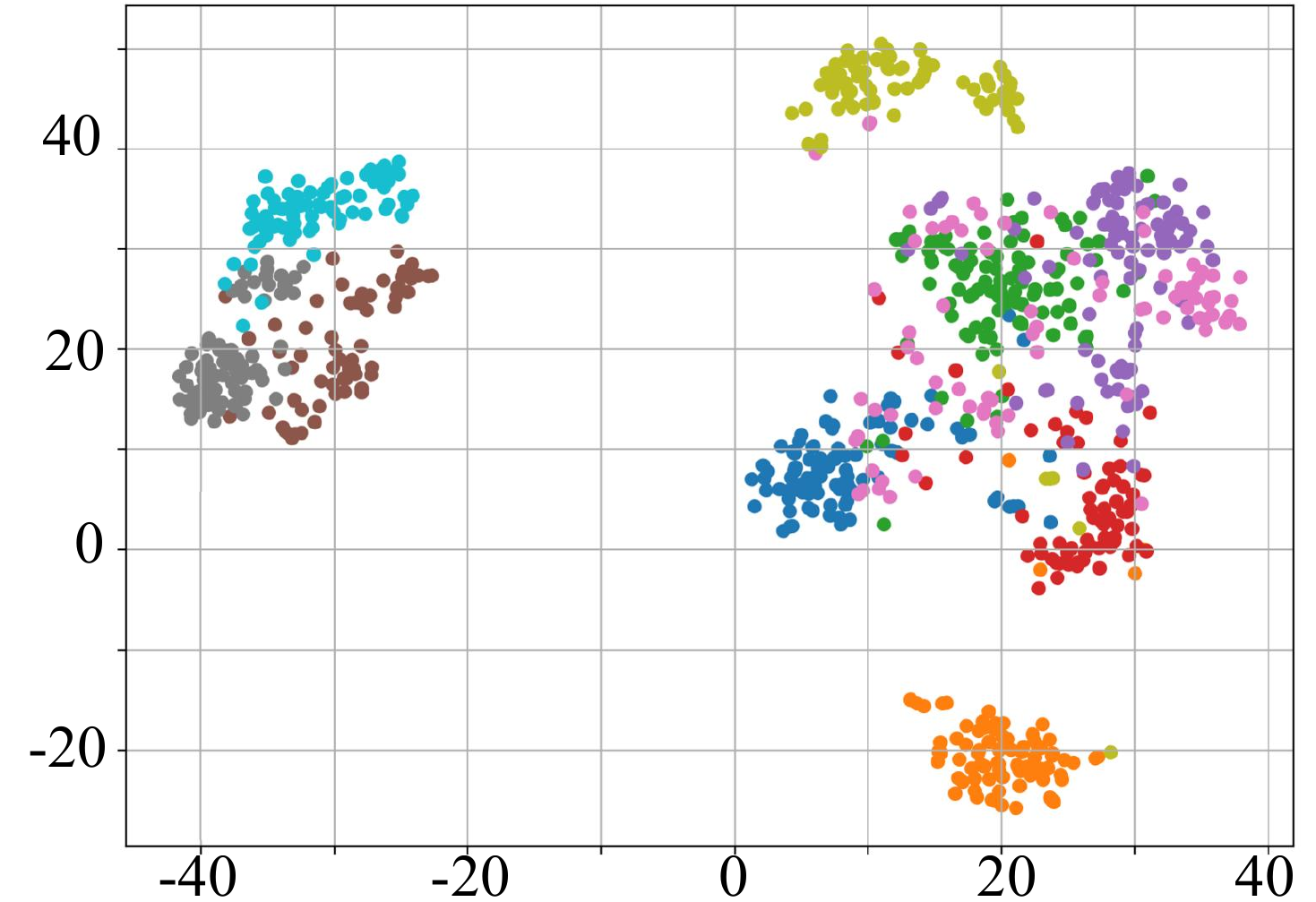}
        \caption{Without STBP}
        \label{fig:c4}
    \end{subfigure}

    \caption{t-SNE visualizations of internal feature representations from the penultimate layer under four training settings: full model and three ablated variants, each missing one learning mechanism. The spatial arrangement reflects how feature distributions vary across configurations.}
    \label{fig:t-SNE}
\end{figure}

We compare four training configurations: the full model incorporating STBP, Hebbian learning, and SBP; and three ablated versions, each omitting one of the mechanisms. The visualization results, shown in Fig.~\ref{fig:t-SNE}, reveal that the full model produces more compact and clearly separated clusters, indicating strong class-wise discrimination and intra-class consistency. In contrast, the ablated variants exhibit more scattered or overlapping feature distributions. These visualizations suggest that the integration of multiple learning rules enriches the network’s representational capacity, supporting better generalization and robustness.
\section{Conclusion}
This paper presents a general multi-neuronal plasticity synergy learning method for spiking neural networks, where diverse plasticity rules—whether biologically inspired or gradient based—can be jointly integrated within a unified architecture. Through systematic design and shared membrane dynamics, these mechanisms interact synergistically without interfering with each other’s learning paths. Extensive experiments on both static and dynamic datasets demonstrate that the proposed framework consistently outperforms existing methods in terms of accuracy and robustness. This method offers a flexible foundation for incorporating additional learning rules and adapting to broader neuromorphic applications, facilitating the development of more versatile and more expressive spiking neural networks.   
\bibliography{aaai2026}

\end{document}